\documentclass[letterpaper, 10 pt, conference]{ieeeconf}  

\IEEEoverridecommandlockouts                              

\usepackage{amsmath} 
\usepackage{graphicx}
\usepackage{multirow}

\title{\LARGE \bf
OmniWheg: An Omnidirectional Wheel-Leg Transformable Robot
}

\author{Ruixiang Cao$^{1}$ Jun Gu$^{1}$ Chen Yu$^{1}$ and Andre Rosendo$^{1}$
\thanks{*This work was not supported by any organization}
\thanks{$^{1}$School of Information Science and Technology,
        ShanghaiTech University, Shanghai, China
        {\tt\small \{caorx, gujun, yuchen, arosendo\}@shanghaitech.edu.cn
        }}%
}

\begin{document}

\maketitle
\thispagestyle{empty}
\pagestyle{empty}

\begin{abstract}
This paper presents the design, analysis, and performance evaluation of an omnidirectional transformable wheel-leg robot called OmniWheg.
%
We design a novel mechanism consisting of a separable omni-wheel and 4-bar linkages, allowing the robot to transform between omni-wheeled and legged modes smoothly. In wheeled mode, the robot can move in all directions and efficiently adjust the relative position of its wheels, while it can overcome common obstacles in legged mode, such as stairs and steps.
%
Unlike other articles studying whegs, this implementation with omnidirectional wheels allows the correction of misalignments between right and left wheels before traversing obstacles, which effectively improves the success rate and simplifies the preparation process before the wheel-leg transformation.
%
We describe the design concept, mechanism, and the dynamic characteristic of the wheel-leg structure. We then evaluate its performance in various scenarios, including passing obstacles, climbing steps of different heights, and turning/moving omnidirectionally. Our results confirm that this mobile platform can overcome common indoor obstacles and move flexibly on the flat ground with the new transformable wheel-leg mechanism, while keeping a high degree of stability.

\end{abstract}


\section{Introduction}

Wheels are the most commonly used way of moving in robot systems because of their high efficiency and simplicity. On the other hand, legged robots that are inspired biologically have shown better performance on rough and complex terrains\cite{mini_cheetah}\cite{MIT_Cheetah_3}. Aiming at combing the advantages of both kinds of mobile platforms, wheel-leg transformable mechanisms are often studied\cite{leg-wheel_module}\cite{chen2013quattroped}\cite{bai2018ldr}\cite{STEP_mobile_platform}. These mechanisms can either actively or passively transform between two modes, wheeled and legged. In wheeled mode, the robot is able to traverse flat surfaces efficiently at a high speed while in legged mode, its ability to overcome obstacles will be significantly improved.

Different designs have been presented on wheel-leg robots. Some of them transform passively between wheeled and legged mode. For example, Wheel Transformer can change the shape of its wheel with a unique passive triggering mechanism to climb over an obstacle 3.25 times its wheel radius\cite{wheel_transformer}. WheeLeR utilizes a geared mechanism, which allows it to transform by simply changing the driving direction\cite{WheeLeR}. Other designs transform their wheels actively by one or more actuators. For example, the STEP robot adopted a 2-DOF five-bar PRRRP mechanism to change the wheels' shape actively\cite{STEP_mobile_platform}. TurboQuad can split its wheel into two separate semi-circle wheels with two geared motors\cite{chen2017turboquad}. These wheel leg robots show a good obstacle-overcoming ability while maintaining the advantages of the wheeled mechanism.

\begin{figure}[t]
  \centering
  \includegraphics[width=\linewidth]{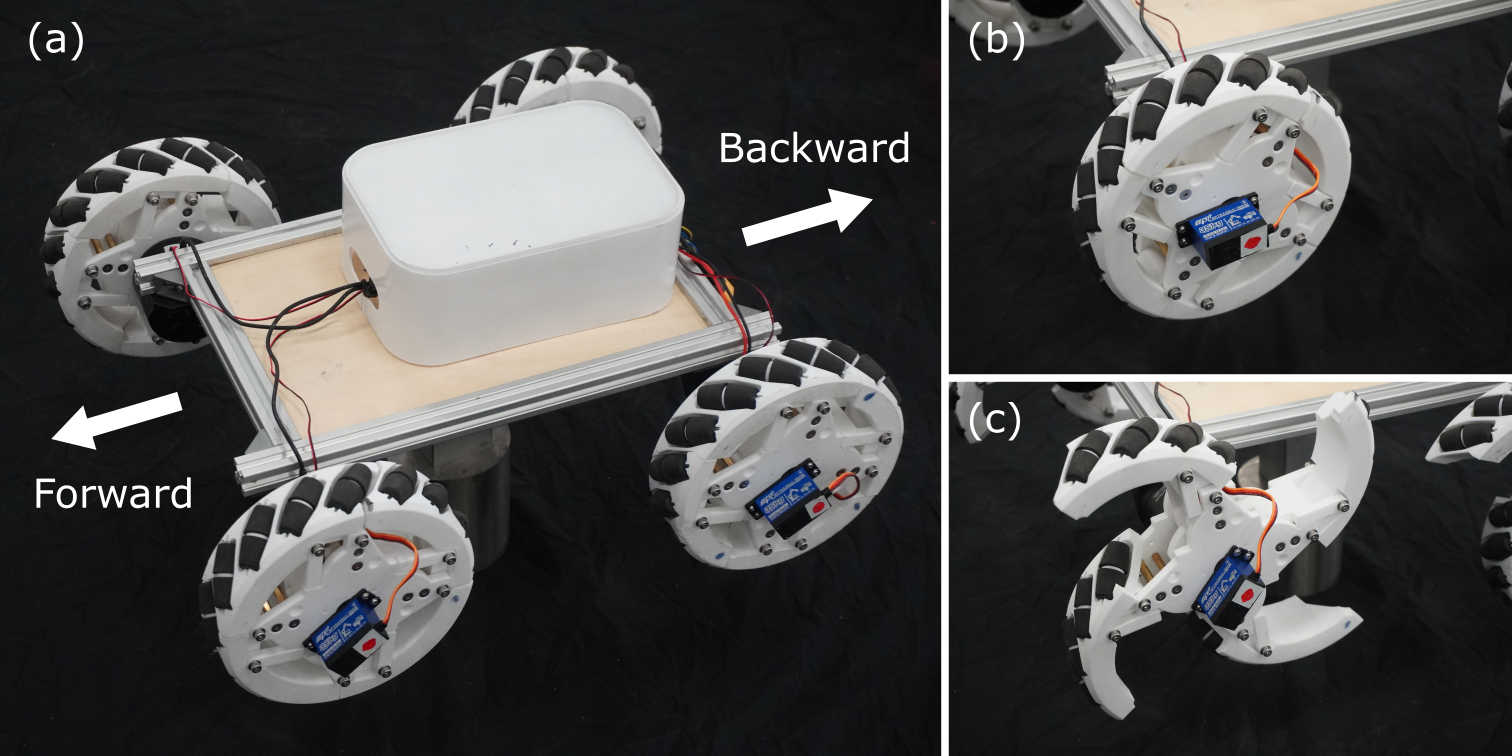}
  \caption{
    \textbf{Overview of OmniWheg}
    (a) The robot is equipped with four omnidirectional transformable wheels.
    (b) Transformable wheels on wheeled mode to move on flat surfaces and align wheels before passing obstacles.
    (c) Transformable wheels on legged mode to pass obstacles.
  }
  \label{fig:overview}
\end{figure}

\begin{figure*}[t]
  \centering
  \includegraphics[width=\linewidth]{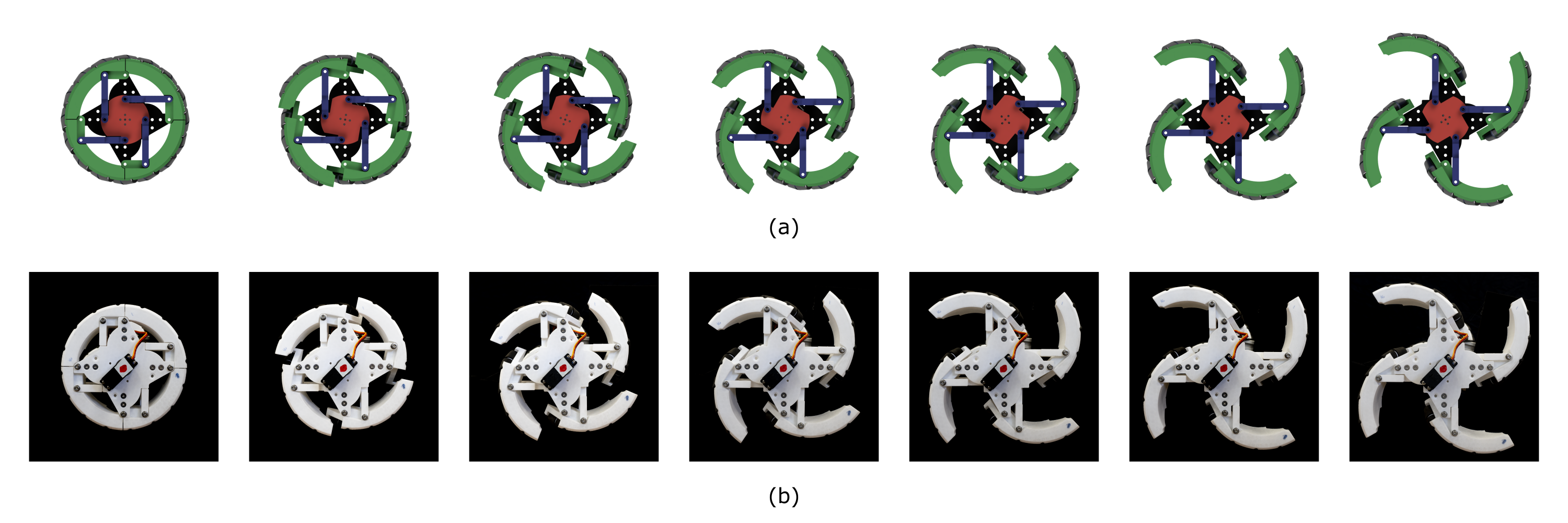}
  \caption{
    \textbf{Unfolding process of the wheel}
    a) Simulated unfolding process of the transformable wheel. The tilting angle of each lobe of the wheel can be adjusted from $0^{\circ}$ to $60^{\circ}$.
    b) Unfolding process of the the real robot. A servo is placed on one side of the wheel to drive the transforming mechanism.
  }
  \label{fig:unfolding}
\end{figure*}

Despite various benefits brought by wheel-leg structures, the alignment of the wheels remains a challenging problem. The misalignment of the two wheels while passing obstacles will adversely affect the success rate or even cause the entire robot to flip over. For example, the Wheel Transformer will tilt when only one of its triggering legs contacts the obstacle before the other one, making the other wheel impossible to transform, thus causing the failure of climbing\cite{wheel_transformer}. In some other wheg robots, for example, the STEP robot, the initial state is fixed and the alignment problem is not discussed in depth\cite{STEP_mobile_platform}.

The adjusting of the relative position between the obstacle and the wheg robot is another important problem. When encountering an obstacle, the position before transforming their wheels is essential for the success of passing over. When the heading direction of the robot is not perpendicular to the obstacle, the possibilities are that its wheels cannot transform simultaneously. LDR robot, for example, when wheels of two sides of the robot don't transform simultaneously because of bad contact position, obstacle-negotiation is possible to fail\cite{bai2018ldr}\cite{bai2021ldr2}. One possible way of solving the problem will be to actively adjust the position according to the shape of the obstacle. However, most existing wheel-legged robots steer by deferentially driving their wheels\cite{wheel_transformer}\cite{bai2018ldr}\cite{irem2020fuhar}, which will inevitably cause a misalignment between left and right wheels, thus making it harder to climb an obstacle stably.

To solve the problem of the misalignment between wheels and the non-ideal relative position between the robot and obstacles, a new method is needed to effectively adjust the pose of the robot and the rotation angle of the wheels while being as compact as possible. Among all different wheel designs, the Mecanum wheel is known for its outstanding ability to maneuver omnidirectionally in narrow spaces\cite{ilon1975wheels}. With a combination of differential wheel motions, the vehicle with Mecanum wheels can move in almost any direction with any rotation. Its discrete structure makes it possible to be applied to transformable wheel-leg designs which commonly require the wheel to be split into several pieces to achieve a better obstacle-overcoming ability.

In our research, a novel mechanism combing the unique advantage of both omnidirectional wheels and the wheel-leg transformable wheel is designed, as shown in Fig.~\ref{fig:overview}. This mechanism utilizes an active multiple 4-bar linkages structure for wheel-leg transforming and a 16-roller omni-wheel design for omnidirectional movement. This mechanism is applied to a robot called OmniWheg. Without manually resetting the initial rotation of wheels and the position of the robot, the robot can climb common indoor obstacles such as steps and stairs with a height up to 2.5 times its wheel radius.

In section II, the design concept and the detailed mechanism of the omnidirectional wheel-leg transformable structure is introduced. The design variables and considerations are also described in this section. Section III introduces the mechatronics and software of the robot. Finally, in section IV, experiments of passing obstacles and maneuvering in all directions are performed using the mobile platform. The result is also discussed. Section V concludes the paper.

\section{Omnidirectional wheel-leg mechanism}

\subsection{Overview}
This section presents the mechanical design and quasi-static analysis of the novel omnidirectional wheel-leg transformable mechanism. This design mainly consists of two parts: A multiple 4-bar linkages structure responsible for transforming the wheel between wheeled and legged modes and the passive rollers surrounding the wheel accountable for enabling the robot to move omnidirectionally.

\subsection{Wheel-leg transforming mechanism}

\begin{figure}[h]
  \centering
  \includegraphics[width=\linewidth]{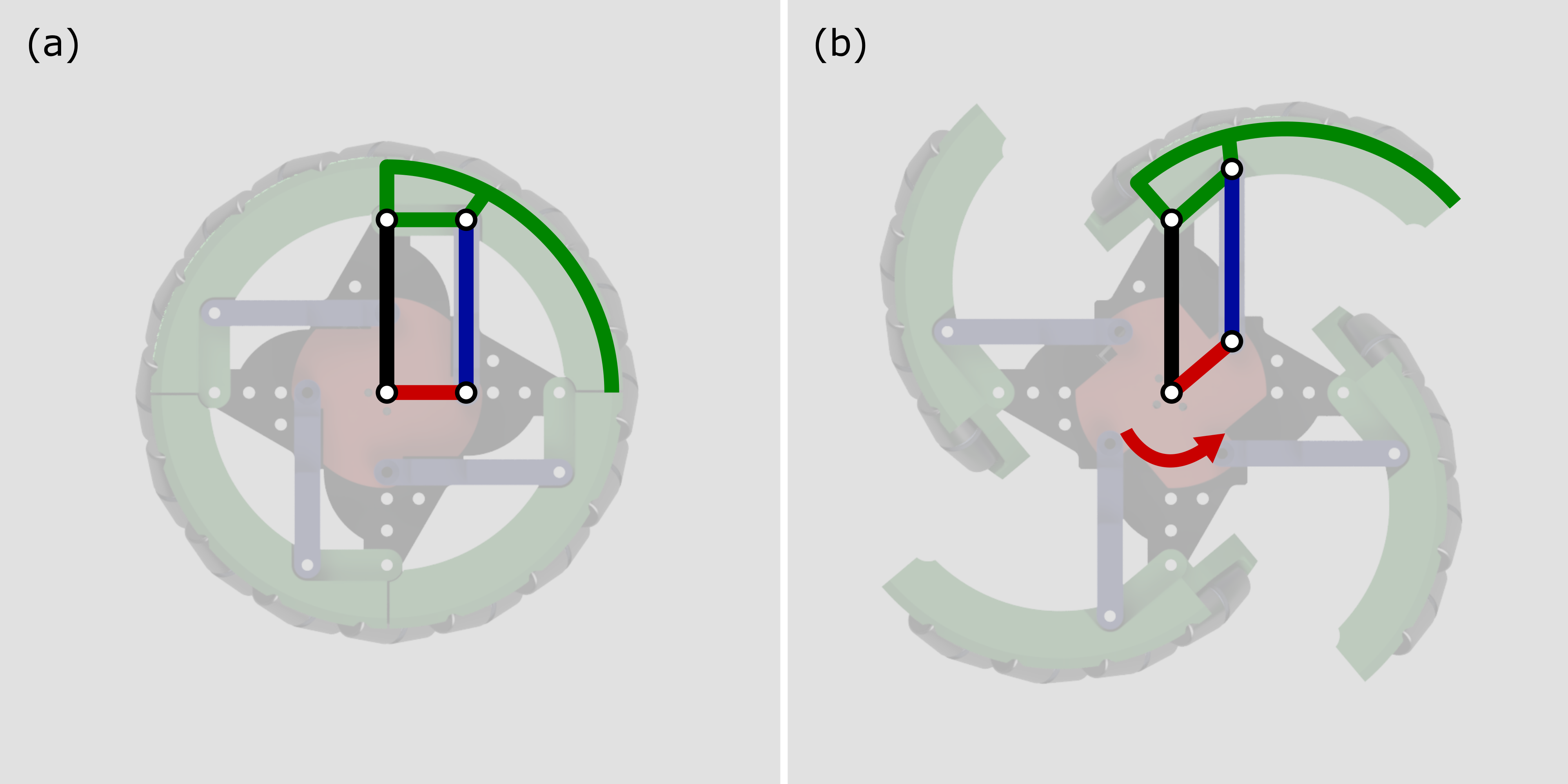}
  \caption{
    \textbf{Four bar linkage mechanism}
    Multiple 4-bar linkages mechanisms are utilized in our design. The mechanism is driven by a servo motor placed in the center of the wheel.
    (a) Four bar linkage mechanism on wheel mode.
    (b) Four bar linkage mechanism on legged mode.
  }
  \label{fig:mechanism}
\end{figure}

The wheel-leg transforming mechanism can be separated into four identical parts. Each part is based on a four-bar linkage mechanism, as shown in Fig.~\ref{fig:mechanism}. These four parts are synchronized by a central disc connected to a servo motor. When this servo motor rotates clockwise, four lobes of the wheel will open synchronously until they are fully expanded, and the robot will be turned into legged mode. When it rotates counterclockwise, four lobes will be retrieved, forming a circular shape, therefore the robot will be turned into wheeled mode. Notice that the rotating direction of the servo motor will be different depending on which position it is installed on the robot.

\subsection{Omnidirectional motion mechanism}

\begin{figure}[h]
  \centering
  \includegraphics[width=\linewidth]{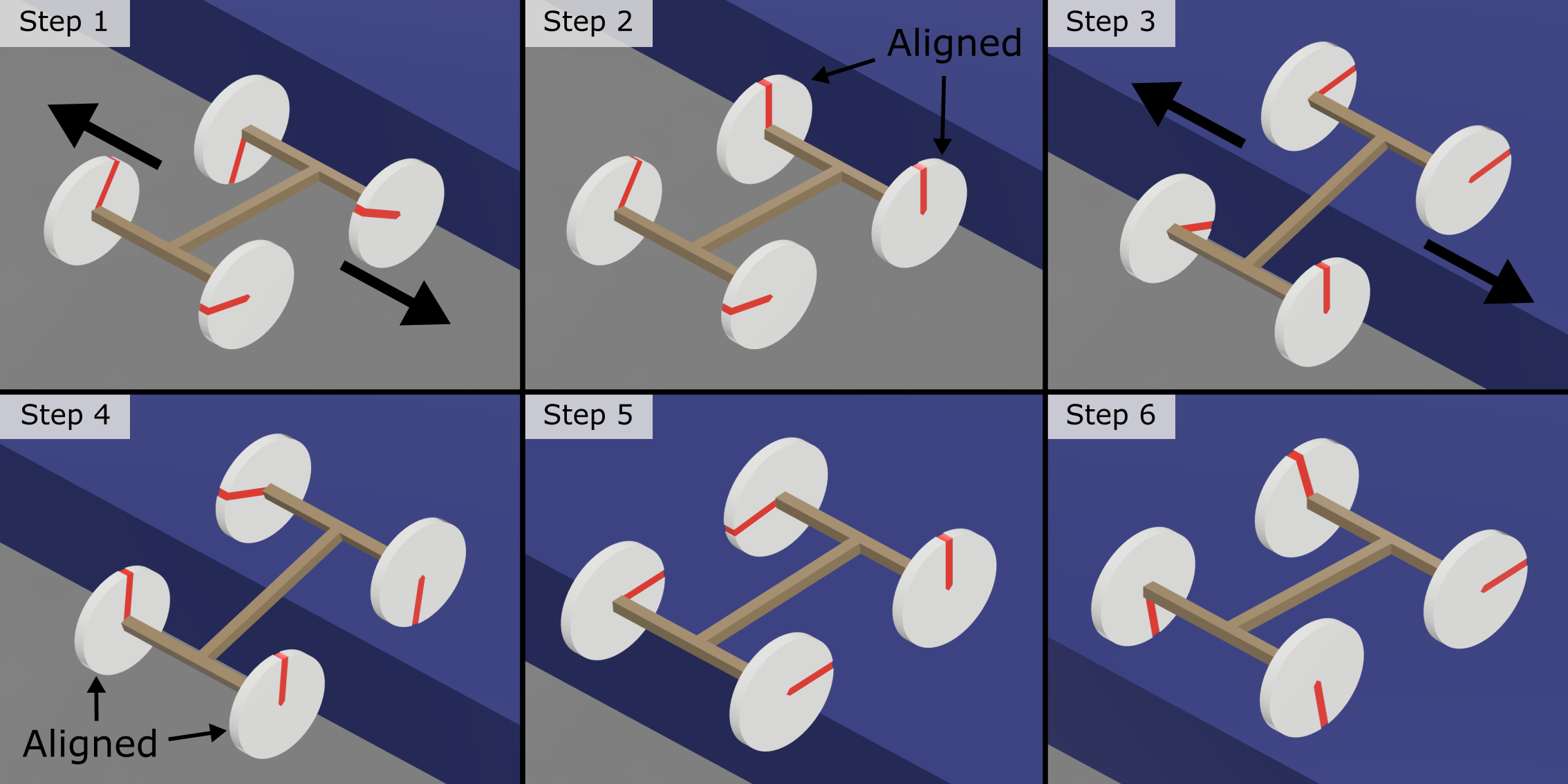}
  \caption{
    \textbf{Motion of wheels during obstacle passing}
    Step 1: Move laterally to align the front wheels.
    Step 2: Two front wheels climb up the obstacle.
    Step 3: Move laterally to align rear wheels.
    Step 4: Two rear wheels climb up the obstacle.
    Step 5: The robot successfully climbed up the obstacle.
  }
  \label{fig:climb}
\end{figure}

A Mecanum  wheel mechanism is applied to our design to enable the robot to move omnidirectionally. This allows the OmniWheg to align its wheels easily before climbing on any obstacles. As shown in Fig.~\ref{fig:climb}, the climbing process can be divided into 6 steps. Step 1: Adjust the robot's relative position to the obstacle so that its heading direction is orthogonal to the obstacle, then moves laterally to align its two front wheels. Step 2: When front wheels are aligned, they are unfolded into legged mode.  Step 3: Front wheels climb up the obstacle in legged mode. After that, they are reset to wheeled mode. Step 4: The robot moves laterally again to align the rear wheels. Step 5: Rear wheels are unfolded into legged mode and climb up the obstacle. Step 6: Rear wheels are reset to wheeled mode, and the robot successfully passed the obstacle.

\begin{figure}[htbp]
  \centering
  \includegraphics[width=\linewidth]{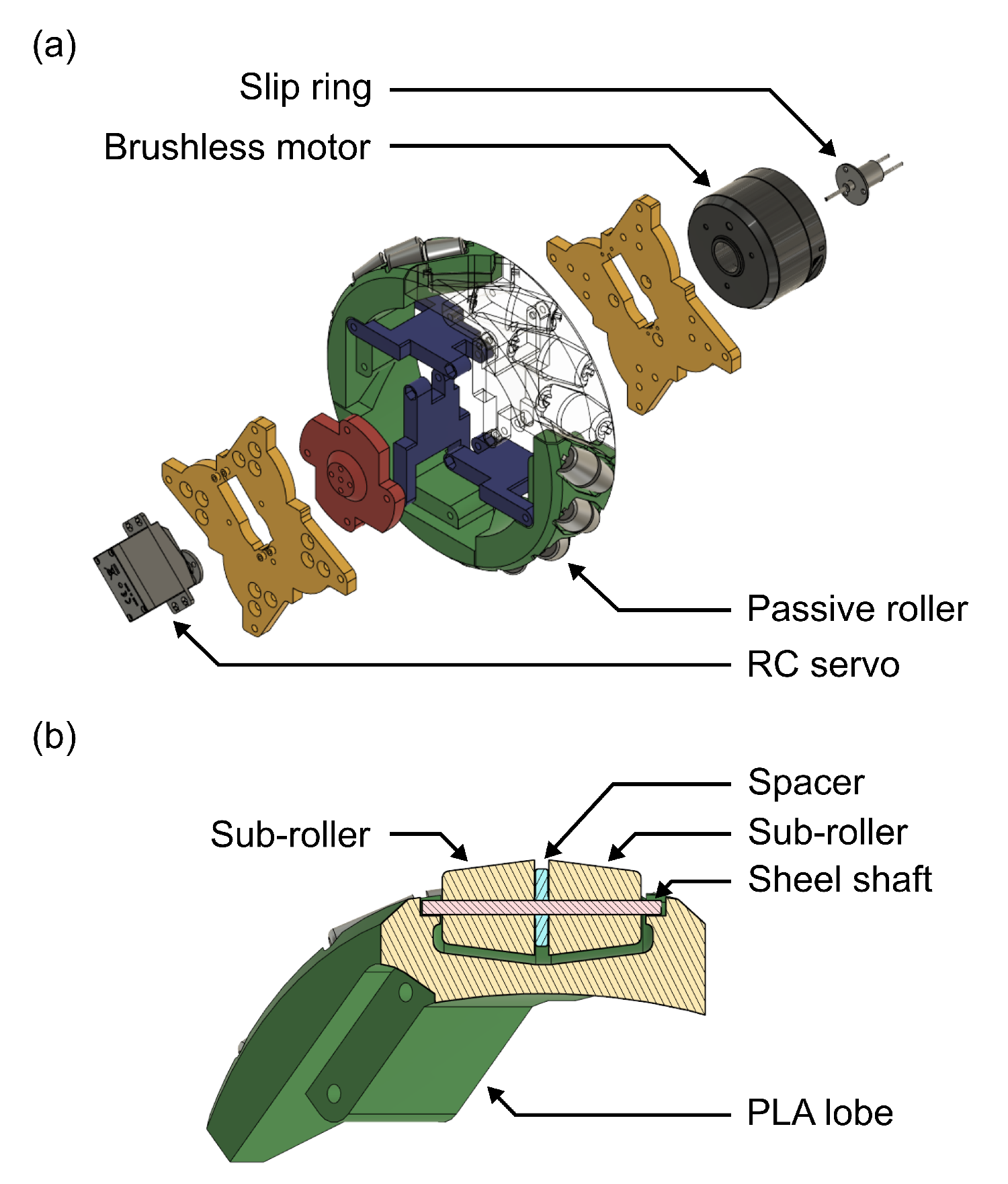}
  \caption{
    \textbf{Structure of the wheel}
    (a) Explosion diagram of one of four transformable wheels. RC servo is used for wheel-leg transformation. A hollow-shaft motor is used for rotating the wheel. A six-channel slip ring is used for transferring power for the servo.
    (b) Passive rollers are mounted on 3D printed lobes.
  }
  \label{fig:explosion}
\end{figure}

The omnidirectional movement is achieved by installing sixteen passive rollers around each wheel of the robot, as shown in Fig.~\ref{fig:explosion}(a). These rollers have an axis of rotation at 45 degrees to the wheel plane and the axle line. By differentially driving four wheels on the robot, the robot can move in almost any direction with any rotation. The rollers can be divided into four groups. Each group of 4 rollers is installed on one lobe of the wheel. When the wheel is opened, the rim of the rollers can be separated without affecting the motion of the transforming mechanism. As shown in Fig.~\ref{fig:explosion}(b), The roller consists of a central metal shaft, two rubber-covered sub-rollers, and a plastic spacer. When the wheel is assembled, the roller will be pushed into one of the four slots on one lobe.

\subsection{Minimum torque requirement for motor}

Four hollow-shaft brushless motors are used for driving the wheels. We perform a quasi-static analysis to find the minimum torque required to climb an obstacle, as shown in Fig.~\ref{fig:quasistatic}(a). The wheel is fully expanded during climbing, and the motor rotates in the CW direction. The required torque on the motor can be acquired by:
\begin{equation}
    \tau_{motor} - F_{wheel} L_1 = \tau_{motor} - F_{wheel} R_{c} \cos{\alpha} > 0 \label{eq:motor_torque}
\end{equation}
When the height of the obstacle is identical to half the maximum radius of the expanded wheel, $\alpha = 0$, the motor torque reaches its maximum. Substituting $F_{wheel} = M_{wheel}g = 13.48(N)$, $R_{c} = 0.132(m)$, $\alpha = 0$ to \eqref{eq:motor_torque}, we will have:
\begin{equation}
    \tau_{motor} > F_{wheel} R_{leg} \cos{\alpha} = 1.779 \ (N \cdot m)
\end{equation}
So the minimum torque required for the motor will be: $\tau_{motor} > 1.779 \ (N \cdot m)$

\subsection{Minimum torque requirement for servo}

\begin{figure}[htbp]
  \centering
  \includegraphics[width=\linewidth]{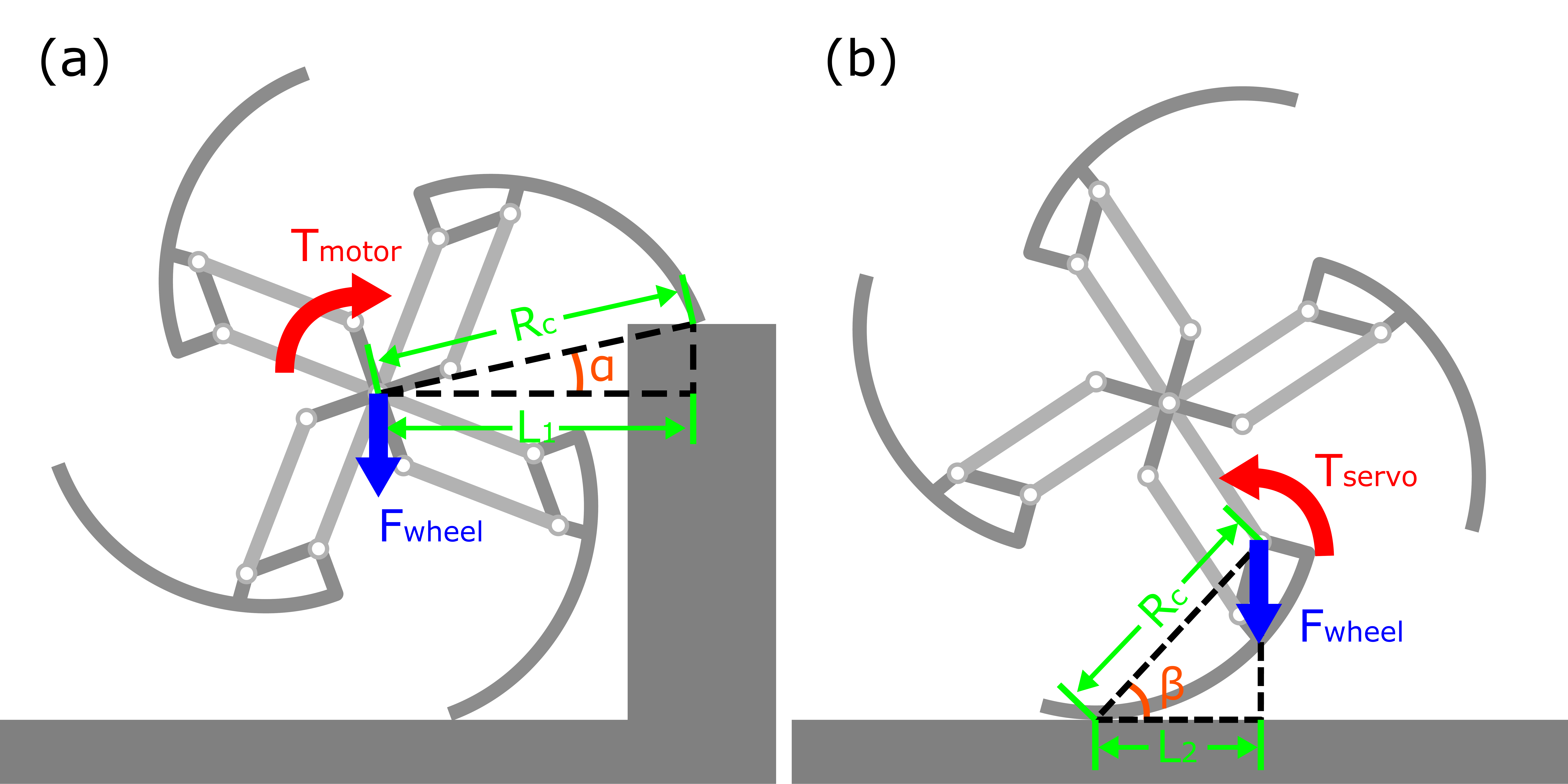}
  \caption{
    \textbf{Quasi-static analysis of the wheel}
    (a) Free body diagram of the wheel passing obstacles.
    (b) Free body diagram of the wheel during wheel unfolding.
  }
  \label{fig:quasistatic}
\end{figure}

The servo is responsible for folding and unfolding the wheel-leg transforming mechanism. This analysis aims to find the minimum torque requirement for the servo to do the transformation successfully. As shown in Fig.~\ref{fig:quasistatic}(b), during the conversion from wheeled to legged mode, the servo spins in the CCW direction, and the torque required on the servo motor is:
\begin{equation}
    \tau_{servo} - F_{wheel} L_2 > 0 \label{eq:servo_torque}
\end{equation}
When the wheel is fully closed, $L_2$ is on its maximum of $65(mm)$, thus $\tau_{servo}$ also reaches its maximum. Substituting $F_{wheel} = M_{wheel} g = 13.48(N)$, $L_2 = 0.065(M)$ to \eqref{eq:servo_torque}, we will have:
\begin{equation}
    \tau_{servo} > F_{wheel} L_2 = 0.876 \ (N \cdot m)
\end{equation}
So the minimum torque required for the servo will be: $\tau_{servo} > 0.876 \ (N \cdot m)$

\begin{table}[htbp]
    \caption{nomenclature}
    \renewcommand\arraystretch{1.2}
    \centering
    \begin{tabular}{|l|l|}
    \hline
        Variable            & Definition                                                \\ \hline
        $R_{wheel}$         & Radius of the wheel in wheeled mode                       \\ \hline
        $R_{leg}$           & Radius of the wheel in legged mode                        \\ \hline
        $\tau_{motor}$      & Torque provided by the brushless motor                    \\ \hline
        $\tau_{servo}$      & Torque provided by the servo motor                        \\ \hline
        $m_{wheel}$         & Weight of the robot performed on the wheel                \\ \hline
        $F_{wheel}$         & Force of the robot weight performed on the wheel          \\ \hline
        $R_{c}$             & Length from the wheel center to the contact point         \\ \hline
        
        $v_{left}$          & Linear velocity of left wheels                            \\ \hline
        $v_{right}$         & Linear velocity of right wheels                           \\ \hline
        $\Delta x$          & Lateral displacement of the robot                         \\ \hline
        $v_x$               & Lateral velocity of the robot                             \\ \hline
        $\Delta \theta$     & Angular difference between left and right wheels in radius  \\ \hline
        $\Delta \theta_{max}$     & Maximum angular displacement for alignment            \\ \hline
        $\Delta \theta_{wheel}$   & Angular displacement for alignment on each wheel      \\ \hline
    \end{tabular}
\end{table}

\subsection{Wheel alignment}

An important advantage of our robot is that it can correct the misalignment of left and right wheels by lateral moving. The required lateral displacement can be calculated. According to the kinematics characteristic of the Mecanum wheel, the lateral velocity of the robot is:

\begin{equation}
    v_x = \left \{
        \begin{aligned}
            v_{left} & = & w_{left} R_{wheel} \\
            - v_{right} & = & - w_{right} R_{wheel} 
        \end{aligned}
    \right.
    \label{eq:velocity}
\end{equation}

The lateral displacement of the robot is:

\begin{equation}
    \Delta x = \Delta \theta R_{wheel}
\end{equation}

Substituting $R_{wheel} = 0.095m$, we will find:

\begin{equation}
    \Delta x = 0.095 \Delta \theta \ (m)
    \label{eq:lateral_displacement}
\end{equation}

Since wheel is consist of four identical lobes, the maximum angular correction needed can be obtained by:

\begin{equation}
    \Delta \theta_{max} = \frac{1}{4} \times 2 \pi \times \frac{1}{2} = \frac{\pi}{4} \ (rad) = 45^\circ
    \label{eq:angular_correction}
\end{equation}

In order to move laterally, two wheels on either side of the robot rotates in opposite direction. As a result, the angular displacement on each wheel is half the overall angular correction needed:

\begin{equation}
    \Delta \theta_{wheel} = \frac{1}{2}  \Delta \theta_{max} = \frac{\pi}{8} \ (rad)
\end{equation}

Combining \eqref{eq:lateral_displacement} and \eqref{eq:angular_correction}, we can find the maximum lateral displacement needed for wheel alignment:

\begin{equation}
    \Delta x = 0.095 \times \frac{\pi}{8} = 0.0373 \ (m)
    \label{eq:displacement}
\end{equation}

According to \eqref{eq:displacement}, the lateral displacement of the robot will be less than 4cm, which is less than tenth of the width of the robot. This shows that the OmniWheg is able to align its wheels with a small space requirement.

\section{Design of the robot}

\subsection{Overall design}

\begin{table}[htbp]
    \caption{Specification of the OmniWheg robot}
    \renewcommand\arraystretch{1.2}
    \centering
    \begin{tabular}{lll}
    \hline
    \multirow{5}{*}{Overall}                     & Weight               & 5.5 $kg$            \\
                                                 & Minimum dimension    & 510$\times$470$\times$250 $mm$   \\
                                                 & Maximum dimension    & 610$\times$470$\times$295 $mm$   \\
                                                 & Controller           & STM32F407           \\
                                                 & Power                & Li-Po battery DC 24V\\
                                                 \hline
    \multirow{5}{*}{Transformable wheel}         & Minimum radius       & 190 $mm$            \\
                                                 & Maximum radius       & 300 $mm$            \\
                                                 & Width                & 80 $mm$             \\
                                                 & Servo motor          & SPT5435LV           \\
                                                 & Drive motor          & GM6020              \\ 
                                                 \hline
    \end{tabular}
    \label{tab:spec}
\end{table}

The overall structure of the omnidirectional transformable mobile platform is shown in Fig.~\ref{fig:overview}. It mainly consists of four omnidirectional transformable wheels, the aluminum frame, and the controlling system.

The frame of the robot is mainly made of aluminum bars due to its light weight and ease of processing. Other parts are mainly 3D printed parts. The support for the motor is 3D printed with a carbon fiber, nylon mixed material. The transformable wheel is also 3D printed with white Polylactic Acid(PLA) material. In order to keep the robot as light as possible, these parts are printed with a fill rate of 15$\%$.

\subsection{Control system}

\begin{figure}[htbp]
  \centering
  \includegraphics[width=\linewidth]{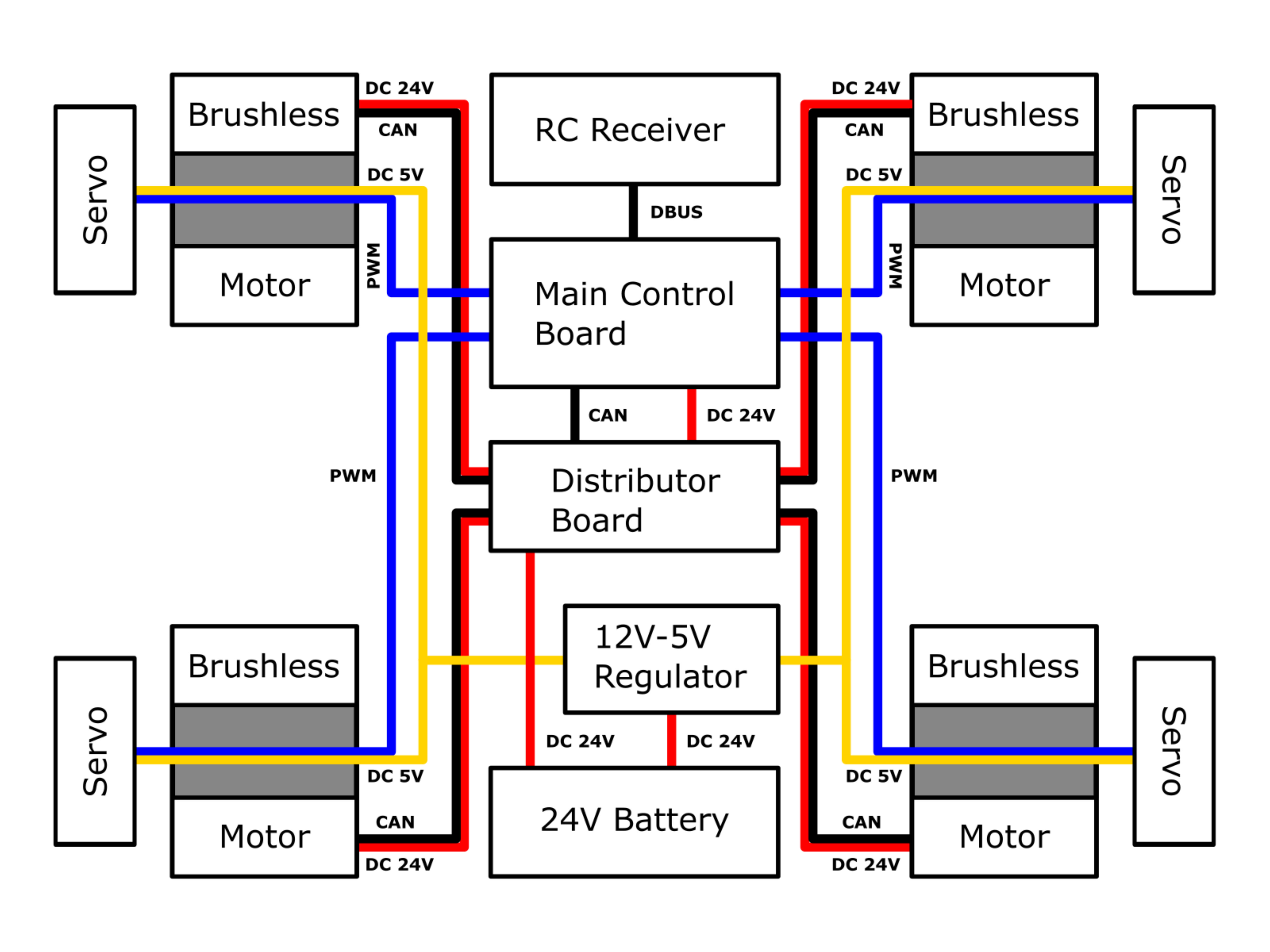}
  \caption{
    \textbf{Electronics of the robot}
    The robot is controlled by a control board with a 32-bit micro processor. Brushless motors are controlled with CAN bus. Servos are controlled with PWM signals.
  }
  \label{fig:electronics}
\end{figure}

To enable the mobile platform to overcome obstacles and move omnidirectionally effectively, we built an electronic control system, as shown in Fig.~\ref{fig:electronics}. The system mainly consists of two layers: the decision layer and the executive layer.

The decision layer is responsible for receiving commands from an RC controller, motion controlling, and generating the required signal for actuators. An STM32F407 micro control unit (MCU) is selected as the central processor because of its rich peripheral and high performance. Two CAN controllers with a bitrate up to 1 Mbps and twelve 16-bit timers are integrated into the MCU for communicating with the actuators. The microprocessor runs at 168MHz, which is adequate for dealing with the omnidirectional motion control algorithm and other high-level environmental perception algorithms to be developed in the future.

The executive layer consists of four Robomaster GM6020 brushless hollow shaft DC motors for driving the wheels and four SPT5435LV  RC servos installed on each wheel for driving the wheel-leg transformation mechanism, as shown in Fig.~\ref{fig:electronics}.  Brushless motors are connected to the control board through a CAN bus running at a bitrate of 1 Mbps, while RC servos are controlled individually by four lines of PWM signals. Since four servos draw around 14A of current at full load, it is impossible to drive them directly with the control board, instead, an individual 5V power supply circuit is used.

\section{Experiment} 

To evaluate the climbing performance of our robot, we conducted 16 experiments for eight different obstacle heights in two different climbing directions. During each experiment, we measured the torque of the front wheels, and the trajectory of the center of the wheel. To find out how much space is needed to align its wheels before climbing the obstacle, We also conducted five additional experiments measuring the robot's lateral displacement to correct the misalignment between right and left wheels.

\subsection{Experiment setup}

An adjustable wooden platform is built as the obstacle. The height of the platform can be adjusted from 10cm to 40cm. The processes of the robot climbing obstacles with a height of 12cm, 14cm, 16cm, 18cm, 20cm, 22cm, 24cm, 26cm were recorded by a camera. Video analyzer Kinovea tracks the center of front wheels from videos. The current of brushless motors is acquired through the CAN messages replied by these motors during the experiment. The torque of these motors are also calculated.

\subsection{Climbing trajectories}

\begin{figure}[htbp]
      \centering
      \includegraphics[width=\linewidth]{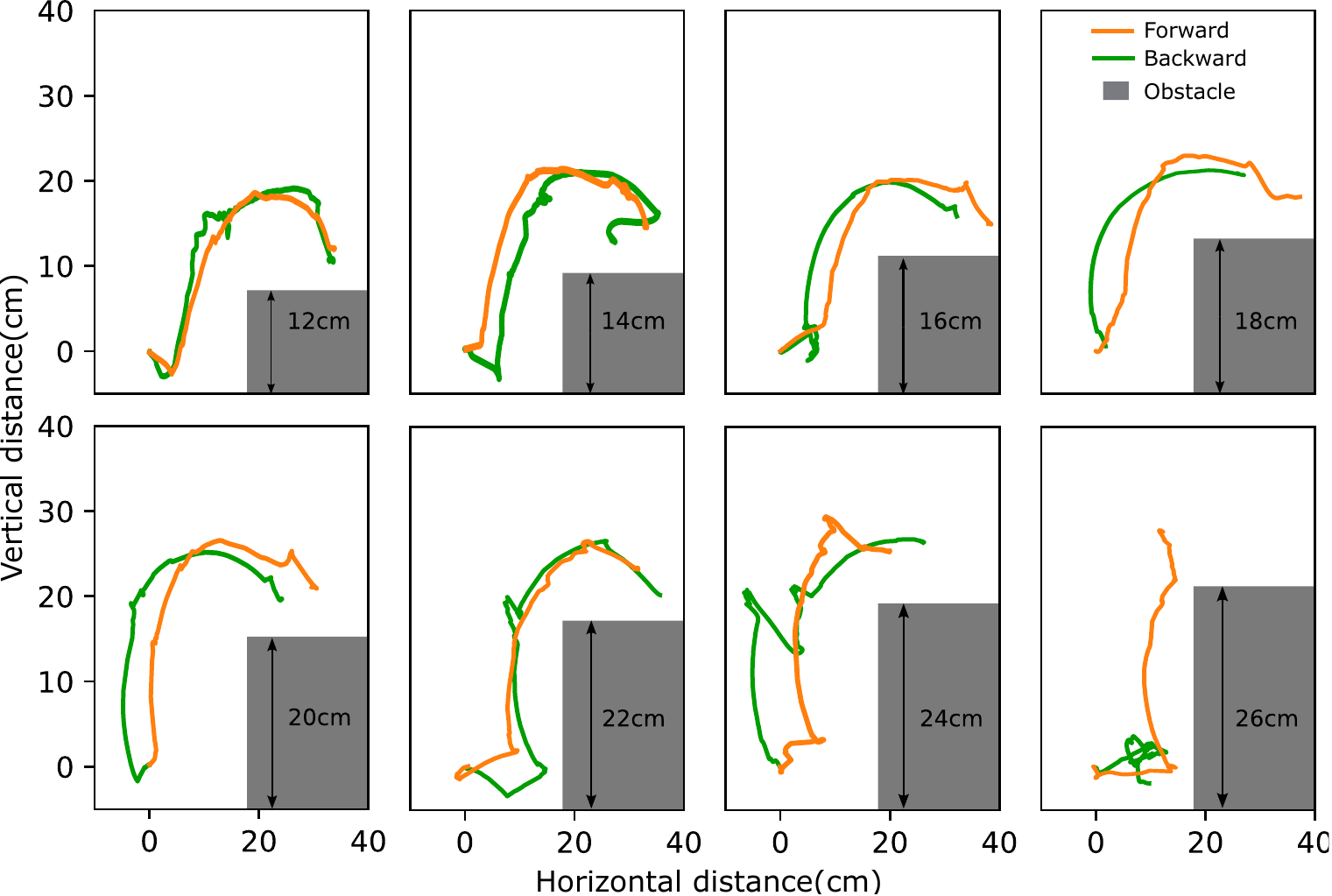}
      \caption{
        \textbf{The trajectories of the wheel center}
        The maximum climbable height is 24cm when moving in the backward direction. The maximum climbable height is 26cm when moving in the forward direction. 
      }
      \label{fig:trajectories}
\end{figure}

To find the maximum height that our robot can climb in both heading directions (forward/backward), eight experiments are performed on each direction. The trajectory of the wheel center is tracked by a camera, as shown in Fig.~\ref{fig:trajectories}. The result shows that the robot can climb the obstacle with height of up to 26cm in the forward direction and 24cm in the backward direction.

By reading the CAN messages replied by brushless motors, we are able to measure the real-time current data of these motors. The torque of them can then be calculated according to the datasheet of the motor:

\begin{equation}
    \tau_{motor} = I_{motor} \times 0.741 N \cdot m / A
\end{equation}

\begin{figure}[htbp]
      \centering
      \includegraphics[width=\linewidth]{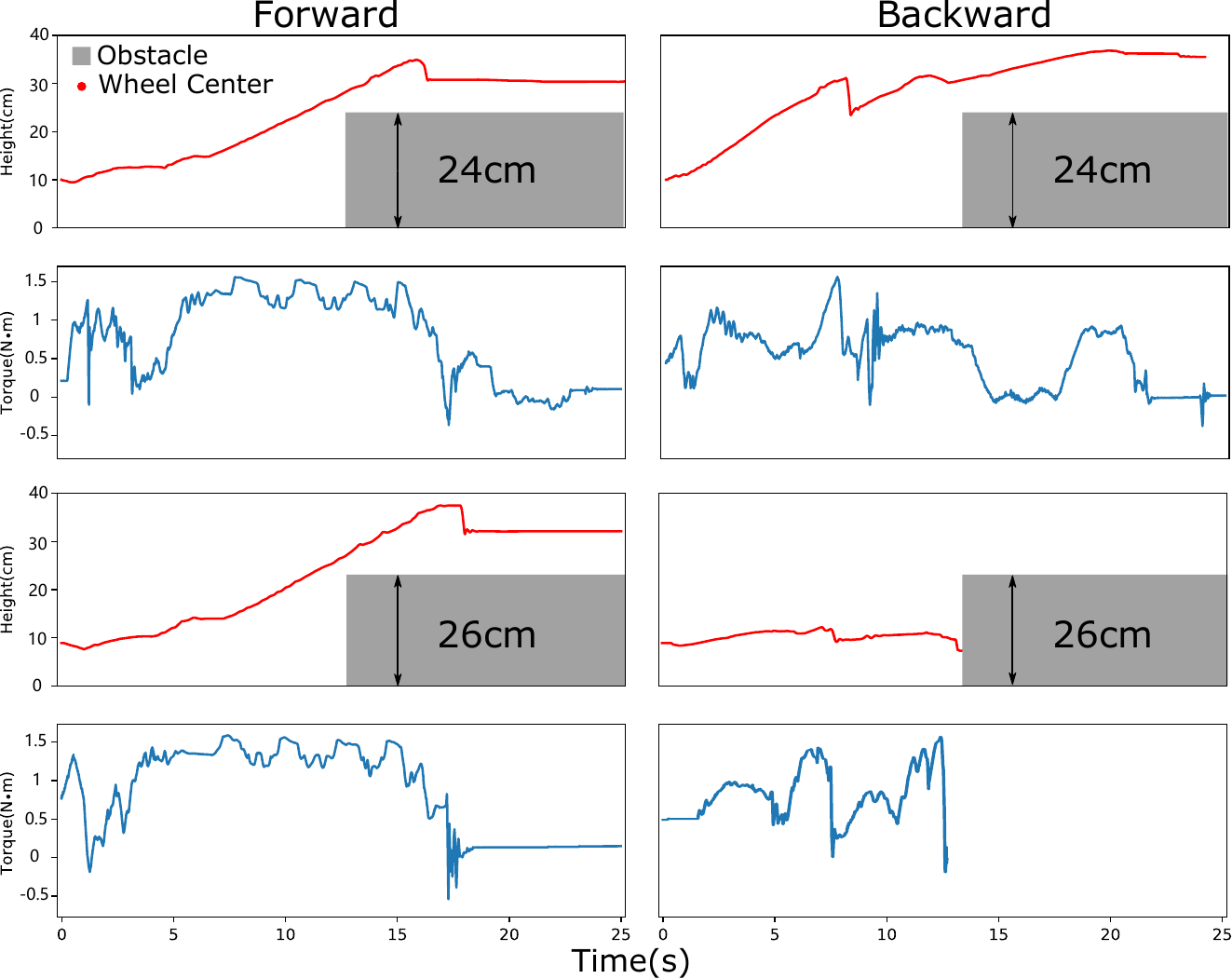}
      \caption{
         \textbf{The torque of the drive motor when climbing over obstacles}
         The torque of the brushless motor when climbing 24cm and 26cm obstacles in two moving directions. On average, the torque and maximum height in the forward direction are higher. The maximum value of torques is about $1.5 (N \cdot m)$. 
      }
      \label{fig:torques}
\end{figure}

The torque data of climbing obstacles with the height of 24cm and 26cm are shown in Fig.~\ref{fig:torques}. The result shows that although the average torque requirement is higher when moving in forward direction, the maximum climbable height is also higher. On the other hand, when climbing in backward direction, the average torque needed is lower, thus a high power efficiency can be acquired.

\subsection{Wheel alignment}

\begin{figure}[htbp]
      \centering
      \includegraphics[width=\linewidth]{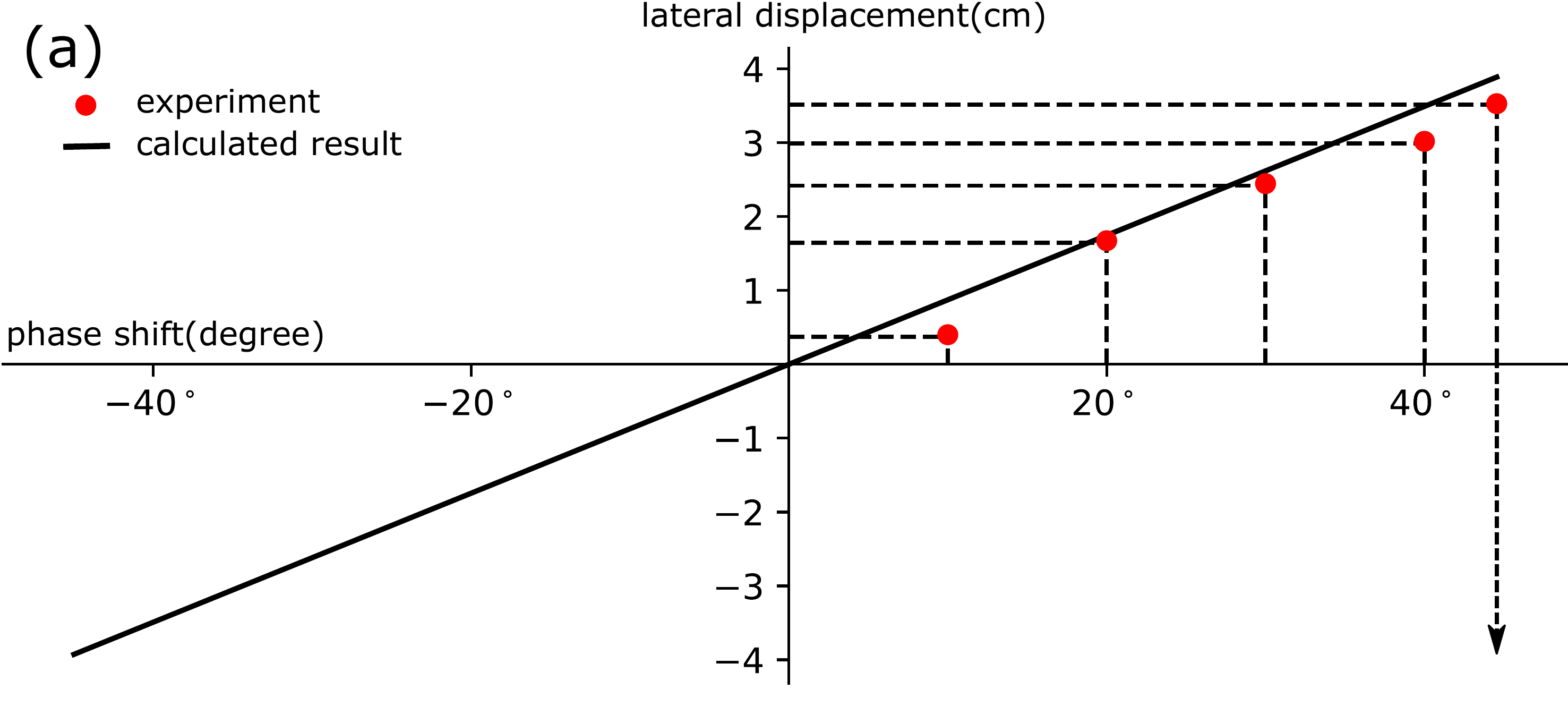}
      \includegraphics[width=\linewidth]{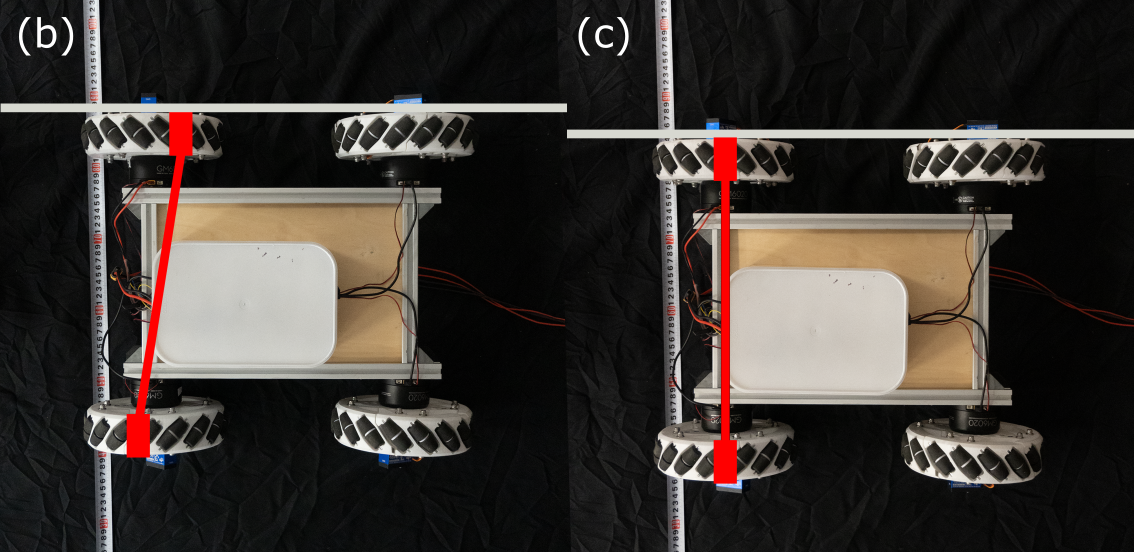}
      \caption{
        \textbf{The relationship between lateral displacement and phase difference of two wheels}\\
        (a) The red points are data points measured by experiments. The phase difference is limited from $-45^{\circ}$ to $45^{\circ}$ due to the central symmetry structure of the wheel. Displacements measured during experiments are smaller than calculated values due to slipping.
        (b) Top view of the OmniWheg before wheel alignment.
        (c) Top view of the OmniWheg after wheel alignment. The lateral displacement of correcting $45^{\circ}$ phase difference is less than 4cm.
      }
      \label{fig:shift}
\end{figure}

To find the maximum lateral displacement required to align wheels in the real world, we analyzed the relationship of the phase difference between wheels and the lateral displacement needed to correct this difference. Notice that only two wheels are tested because of the symmetrical design of the robot. We have calculated this relationship in \eqref{eq:lateral_displacement}. The result of the experiment and the calculated result is shown together in Fig.~\ref{fig:shift}. It shows that the lateral displacement in the real world is lower than the calculated result. This is because of the slip between the wheel and the ground surface. The maximum displacement required is less than 4cm. This confirms that OmniWheg only needs a small clearance to align its wheels.

\section{Conclusion}

In this work, we presented a novel omnidirectional wheel-leg transformable mechanism and its implementation on a robot called OmniWheg. The robot can transform between omni-wheeled and legged mode to traverse both flat surfaces and common indoor obstacles. Compared to other wheel-leg transformable mechanisms, this mechanism enables the wheel-leg robot to easily adjust its relative position to the obstacle before passing it and align its wheels efficiently during the climbing process.

Quasi-static analysis are conducted to acquire the minimum requirement of torque on both the servo motor for transforming and the brushless motor for rotating the wheel.

Experiments on climbing obstacles of different heights are conducted. Our results indicate that the robot can climb obstacles with the height of up to 26cm in the forward climbing direction and 24cm in the backward direction. The maximum space needed for aligning its wheels is found to be 3.7cm, which facilitates the usage of this robot on everyday applications.

In our future works, we plan to further improve our design: The hollow-shaft brushless motor can be replaced by a stronger gear motor and a timing belt to reach a higher climbing ability. Rubber or silicon cover can be added to the wheel lobe at its contacting point with the obstacle, which will reduce the chance of slipping during the climbing process. A depth sensor, ultra sonic sensor and a powerful computer can be installed on the robot as the perception layer, so that it can navigate in a unknown environment.

\bibliographystyle{IEEEtran}
\bibliography{IEEEabrv,IEEEexample}

\end{document}